\documentclass[10pt,conference]{IEEEtran}

\usepackage[T1]{fontenc}
\usepackage[utf8]{inputenc}
\usepackage{lmodern}
\usepackage{microtype}
\usepackage{amsmath,amssymb}
\usepackage{booktabs}
\usepackage{multirow}
\usepackage{array}
\usepackage{graphicx}
\usepackage{xcolor}
\usepackage{url}
\usepackage[hidelinks]{hyperref}
\usepackage{enumitem}
\usepackage{listings}
\usepackage{tikz}
\usepackage{balance}
\usetikzlibrary{arrows.meta,positioning,fit,calc}

\definecolor{accent}{RGB}{36,82,126}
\definecolor{softblue}{RGB}{235,243,250}
\definecolor{softgray}{RGB}{246,247,249}
\definecolor{codegray}{RGB}{70,70,70}

\hypersetup{
  pdftitle={Zhengrong-Spatial-Aware 1.0: Advancing 3D Multimodal Intelligence},
  pdfauthor={Yu Zhang, Xiaogang Wang, Yuhang Cheng, Jingrui Li, Wenbo Chen},
  colorlinks=true,
  linkcolor=accent,
  citecolor=accent,
  urlcolor=accent
}

\setlist[itemize]{leftmargin=*,nosep}
\setlist[enumerate]{leftmargin=*,nosep}

\lstdefinestyle{tech}{
  basicstyle=\ttfamily\scriptsize,
  backgroundcolor=\color{softgray},
  frame=single,
  rulecolor=\color{black!15},
  breaklines=true,
  columns=fullflexible,
  showstringspaces=false,
  keywordstyle=\color{accent}\bfseries,
  commentstyle=\color{codegray},
  xleftmargin=2pt,
  xrightmargin=2pt
}

\title{Zhengrong-Spatial-Aware 1.0:\\
Advancing 3D Multimodal Intelligence}

\author{
\IEEEauthorblockN{Yu Zhang\textsuperscript{1}, Xiaogang Wang\textsuperscript{2}, Yuhang Cheng\textsuperscript{1}, Jingrui Li\textsuperscript{1}, Wenbo Chen\textsuperscript{1}}
\IEEEauthorblockA{\textsuperscript{1}Beijing Freedo Technology Co., Ltd.\\
\textsuperscript{2}Southwest University}
}

\begin{document}
\maketitle

\begin{abstract}
This report presents the implementation and validation of two end-to-end multimodal adaptation pipelines for 3D-aware visual question answering on a custom spatial dataset. The first pipeline fine-tunes HiSpatial-3B, which augments RGB observations with metric-scale 3D point maps; the second adapts Qwen3-VL through memory-efficient QLoRA-based supervised fine-tuning. The completed engineering stack covers dataset conversion, image--annotation alignment, monocular geometry preprocessing, instruction construction, mixed-domain training, checkpoint management, batch inference, and automated evaluation. On the target spatial benchmark, HiSpatial adaptation increases overall accuracy from 1.73\% before fine-tuning to 79.89\% at 3,600 steps. On a separately collected styled split, accuracy improves from 14.9\% to 55.1\%, indicating transfer beyond the original training set. The Qwen3-VL pipeline is likewise implemented end-to-end for both 8B- and 32B-scale models using mixed-domain QLoRA adaptation. The resulting accuracy improves from 20.74\% to 45.27\% for Qwen3-VL-8B and from 30.79\% to 47.60\% for Qwen3-VL-32B. Together, these results establish two complementary and reproducible adaptation pipelines for geometry-aware and general-purpose multimodal spatial reasoning. These results establish a reproducible engineering foundation for subsequent work on metric 3D reasoning, spatial grounding, and domain-adapted multimodal systems, while also identifying training-data composition as a central factor in preserving general multimodal capability during specialization.
\end{abstract}


\section{Introduction}
Vision-language models (VLMs) have achieved strong performance in semantic recognition, captioning, and visual question answering, but reliable metric 3D reasoning remains substantially more challenging when the model observes only a 2D image. The target project focuses on spatial tasks that require both semantic understanding and geometric sensitivity, including object counting, building-height estimation, footprint comparison, pairwise distance estimation, spatial ordering, and relative-direction reasoning.

The objective of this work is to build and validate a practical adaptation stack for strong open-source multimodal models on the provided 3D-aware dataset, rather than to train a new foundation model from scratch. Two complementary model families are studied. HiSpatial explicitly injects geometric information through a metric point-map branch and is therefore naturally suited to quantitative and relational spatial reasoning. Qwen3-VL, by contrast, provides a strong general-purpose multimodal baseline with learned spatial and 3D grounding capabilities, enabling us to evaluate how far task specialization can be achieved through instruction tuning alone.

The completed work delivers the following engineering outcomes:
\begin{itemize}
    \item a model-independent data layer that converts the provided scenes into aligned image--question--answer records and supports reproducible train/test splits;
    \item an RGB+3D preprocessing pipeline for HiSpatial based on monocular point-map estimation and WebDataset serialization;
    \item mixed-domain HiSpatial supervised fine-tuning, checkpointed inference, and automated evaluation, with target accuracy improving from 1.73\% to 79.89\%;
    \item Qwen3-VL QLoRA pipelines for both 8B and 32B variants, including instruction conversion, low-memory training, checkpoint loading, and standardized evaluation;
    \item a unified prediction/evaluation interface for comparing spatial VLMs on the same JSONL benchmark; and
    \item an empirical diagnosis of narrow-domain over-specialization in Qwen3-VL, which motivates mixed-domain sampling and stronger retention controls in subsequent experiments.
\end{itemize}

\section{Model Principles and Literature Background}
\subsection{HiSpatial: Hierarchical 3D Spatial Understanding}
HiSpatial formulates 3D spatial intelligence as a four-level hierarchy: basic geometric perception, object-level spatial understanding, inter-object relational understanding, and abstract spatial reasoning~\cite{hispatial}. The hierarchy spans low-level depth or point queries, object-centric properties such as physical size and orientation, pairwise relations such as distance and direction, and higher-level tasks including perspective taking and multi-step spatial problem solving.

A central architectural choice is to supplement RGB appearance with explicit geometric information. During both training and inference, a monocular geometry estimator produces a point map
\begin{equation}
    \mathbf{P}=G(\mathbf{I}), \qquad \mathbf{P}\in\mathbb{R}^{H\times W\times 3},
\end{equation}
where $\mathbf{I}$ denotes the RGB image and each pixel is associated with a 3D coordinate. HiSpatial uses MoGe-style monocular geometry estimation, whose original formulation predicts an affine-invariant point map from a single image~\cite{moge}. Metric-scale point maps are especially relevant to the target benchmark because several tasks require quantitative reasoning in physical units rather than only qualitative spatial relations.

HiSpatial-3B is built on a PaliGemma2-3B visual-language backbone. PaliGemma 2 combines a SigLIP vision encoder with the Gemma 2 language model and is explicitly designed as a transfer-friendly VLM family~\cite{paligemma2}. HiSpatial adds 3D-aware processing so that visual representations can be conditioned on both appearance and geometry. A simplified representation of the learning objective is
\begin{align}
    \mathbf{v} &= E_{\text{rgb}}(\mathbf{I}), \\
    \mathbf{s} &= E_{\text{xyz}}(\mathbf{P}), \\
    \mathbf{h} &= F(\mathbf{v},\mathbf{s},\mathbf{q}), \\
    \mathcal{L}_{\text{SFT}} &= -\sum_{t}\log p_{\theta}(y_t\mid y_{<t},\mathbf{h}),
\end{align}
where $\mathbf{q}$ is a spatial question and $y_t$ is the target answer token. This formulation is well matched to the project dataset, in which supervision is defined by physically meaningful geometric quantities and relations rather than by object semantics alone.

\subsection{Qwen3-VL: General Multimodal Reasoning with 3D Grounding}
Qwen3-VL follows a three-module multimodal architecture comprising a vision encoder, an MLP-based vision--language merger, and a Qwen3 language-model decoder~\cite{qwen3vl}. Relative to earlier Qwen-VL generations, the model introduces interleaved MRoPE for improved spatial-temporal position modeling and DeepStack for injecting multi-level visual features into the language model. Its training recipe also explicitly includes spatial reasoning and 3D grounding data, making it a suitable general-purpose counterpart to the more geometry-specialized HiSpatial pipeline.

For adaptation in this project, Qwen3-VL is treated as a general multimodal foundation model and fine-tuned with a QLoRA-style recipe. QLoRA keeps the pretrained base model in low precision while optimizing low-rank adapters~\cite{qlora}. For a frozen quantized weight matrix $\widehat{\mathbf{W}}_0$, the adapted transformation is
\begin{equation}
    \mathbf{y}=\left(\widehat{\mathbf{W}}_0 + \alpha\mathbf{B}\mathbf{A}\right)\mathbf{x},
\end{equation}
where $\mathbf{A}$ and $\mathbf{B}$ are low-rank trainable matrices. This substantially reduces memory use and makes 8B/32B adaptation feasible on limited GPU resources.

The two branches provide complementary adaptation strategies. HiSpatial introduces explicit geometry at the input level, whereas Qwen3-VL relies primarily on learned visual-language representations and spatial priors. Evaluating both pipelines therefore helps separate limitations in geometric reconstruction from limitations introduced by domain adaptation, instruction formatting, or language-side reasoning.

\begin{figure*}[t]
    \centering
    \includegraphics[width=0.95\textwidth]{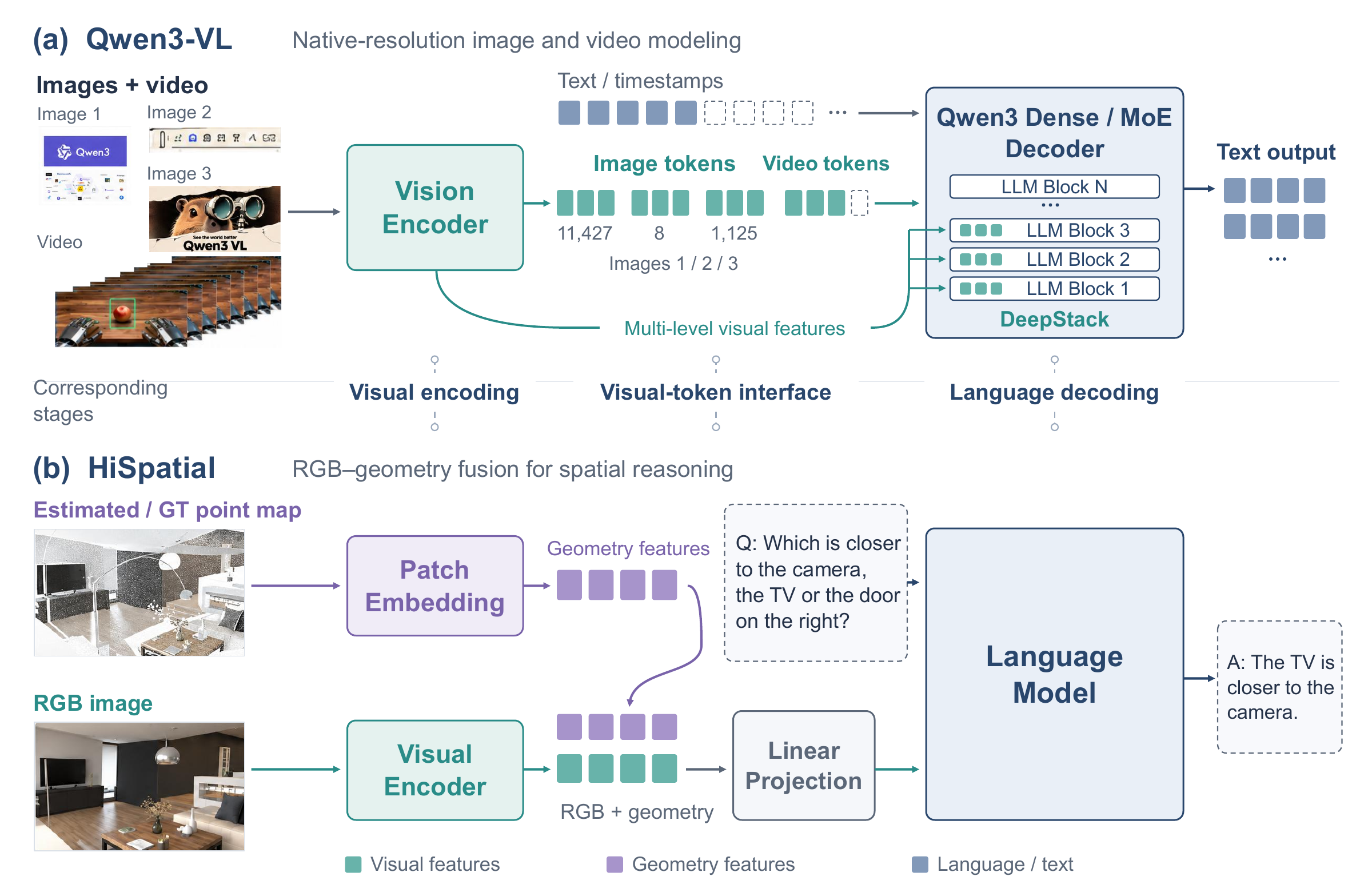}
    \caption{Reference model architectures used in this project. The HiSpatial architecture integrates RGB appearance with a metric-scale point map for geometry-aware spatial reasoning, while Qwen3-VL employs a vision encoder, vision--language merger, and Qwen3 decoder with DeepStack-based multi-level visual feature injection~\cite{hispatial,qwen3vl}.}
    \label{fig:model_arch}
\end{figure*}

\section{Provided Dataset and Task Formulation}
\subsection{Dataset Organization}
The provided dataset was collected and refined in several batches. The implemented pipeline supports raw and styled RGB images, colored bounding-box overlays, instance masks, camera-pose metadata, per-scene annotations, and JSONL-formatted VQA supervision. For the building-centric scenes, CAD/GLB-derived 3D assets are also available and can be incorporated in future geometry-aware extensions.

A typical VQA item is represented as
\begin{lstlisting}[style=tech,language={}]
{
  "sample_id": "sample_000123",
  "q_idx": "q7",
  "question": "What is the distance between ...?",
  "gt": "The horizontal center distance is ... meters."
}
\end{lstlisting}
Each sample identifier is deterministically mapped to its corresponding visual input, allowing the same metadata to be reused consistently for training, batch inference, and checkpoint evaluation.

\subsection{Spatial Question Types}
The benchmark contains twelve question categories. Table~\ref{tab:tasks} groups them by the underlying capability being evaluated rather than by their internal question index.

\begin{table}[t]
\centering
\caption{Target spatial VQA tasks in the provided dataset.}
\label{tab:tasks}
\small
\begin{tabular}{p{0.30\columnwidth} p{0.62\columnwidth}}
\toprule
\textbf{Capability} & \textbf{Representative questions} \\
\midrule
Counting & number of buildings in a scene \\
Metric estimation & building height; pairwise center distance \\
Extremum reasoning & tallest/lowest building; largest/smallest footprint \\
Ordering & footprint ranking; left-to-right and top-to-bottom order \\
Relational reasoning & farthest pair; relative spatial direction \\
\bottomrule
\end{tabular}
\end{table}

Evaluation is task-aware. Counting questions use direct comparison, metric questions such as height and distance use numerical parsing with a relative-tolerance criterion, and open-form relational answers are normalized and evaluated semantically. This design makes the evaluator robust to surface-form variation while preserving stricter treatment of quantitative outputs.

\begin{figure}[t]
\centering
\IfFileExists{data4_color_000000.png}{%
  \includegraphics[width=0.98\columnwidth]{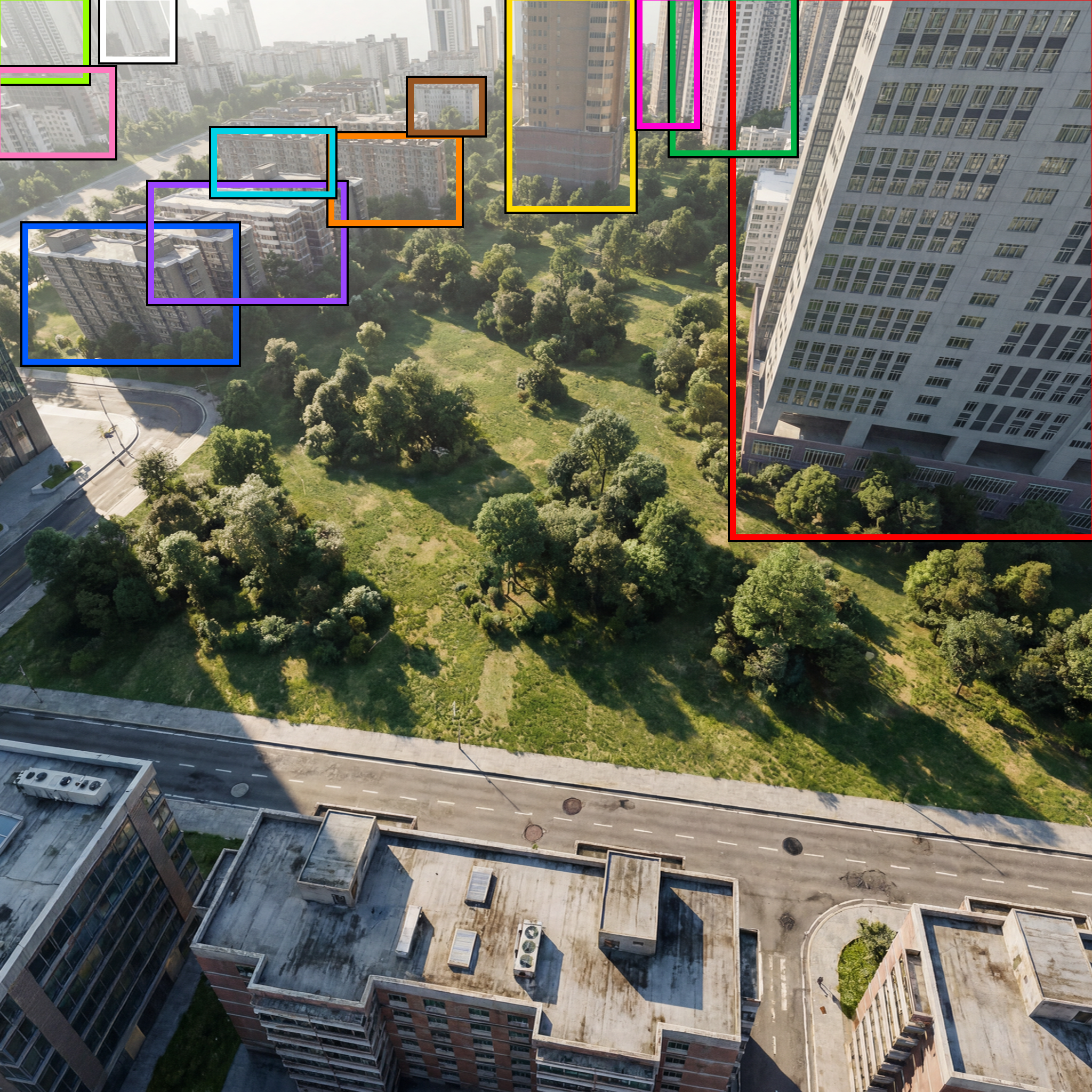}
}{%
  \fbox{\parbox[c][3.2cm][c]{0.92\columnwidth}{\centering
  Dataset example placeholder.\\[3pt]
  Place \texttt{data4\_color\_000000.png} beside this file to render the provided sample.}}
}
\caption{Example visual input from the provided spatial dataset. The engineering pipeline can consume original RGB images as well as styled or region-highlighted variants.}
\label{fig:data}
\end{figure}

\section{Engineering Pipeline}
\subsection{End-to-End System Overview}
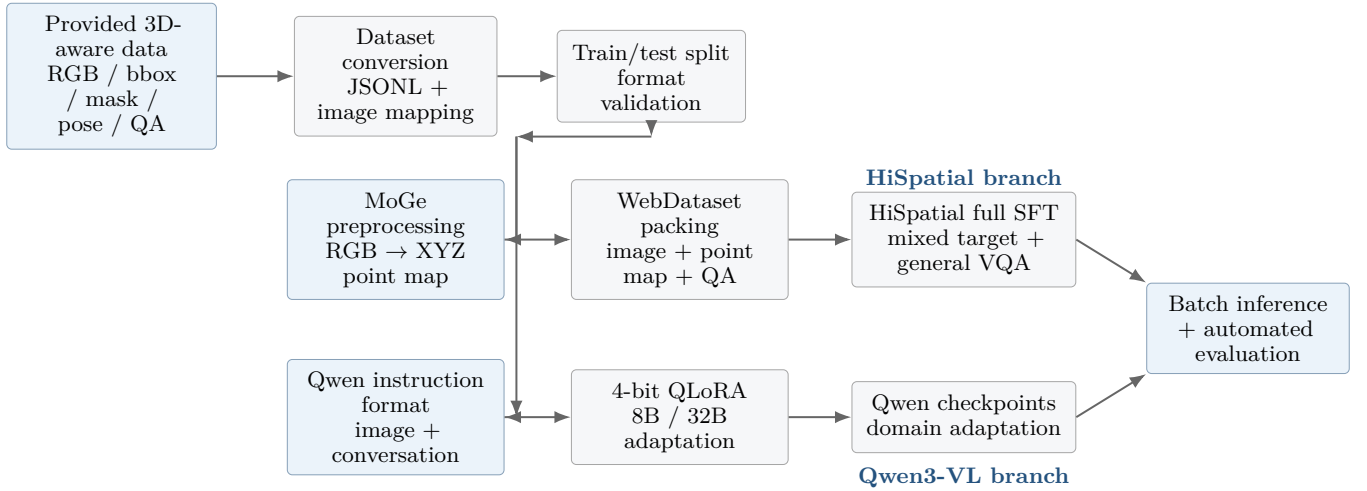
\begin{figure*}[t]
\centering
\resizebox{0.98\textwidth}{!}{%
\begin{tikzpicture}[
  font=\small,
  box/.style={draw=black!35, rounded corners=2pt, fill=softgray, minimum height=10mm, align=center, inner sep=5pt},
  bluebox/.style={box, fill=softblue, draw=accent!55},
  arr/.style={-Latex, thick, draw=black!60}
]
% Shared preprocessing trunk
\node[bluebox, text width=2.6cm] (raw) at (0,0) {Provided 3D-aware data\\RGB / bbox / mask / pose / QA};
\node[box, text width=2.5cm] (convert) at (4.0,0) {Dataset conversion\\JSONL + image mapping};
\node[box, text width=2.3cm] (split) at (7.6,0) {Train/test split\\format validation};

% HiSpatial branch
\node[bluebox, text width=2.7cm] (moge) at (4.0,-2.3) {MoGe preprocessing\\RGB $\rightarrow$ XYZ point map};
\node[box, text width=2.7cm] (webdata) at (8.0,-2.3) {WebDataset packing\\image + point map + QA};
\node[box, text width=2.8cm] (htrain) at (12.0,-2.3) {HiSpatial full SFT\\mixed target + general VQA};

% Qwen branch
\node[bluebox, text width=2.7cm] (qfmt) at (4.0,-4.8) {Qwen instruction format\\image + conversation};
\node[box, text width=2.7cm] (qlora) at (8.0,-4.8) {4-bit QLoRA\\8B / 32B adaptation};
\node[box, text width=2.8cm] (qtrain) at (12.0,-4.8) {Qwen checkpoints\\domain adaptation};

% Shared evaluation
\node[bluebox, text width=2.5cm] (eval) at (16.0,-3.55) {Batch inference\\+ automated evaluation};

% Trunk arrows
\draw[arr] (raw.east) -- (convert.west);
\draw[arr] (convert.east) -- (split.west);

% Clean branch split: route both branches through explicit fork points
\coordinate (fork1) at (7.6,-0.85);
\coordinate (fork2) at (5.7,-0.85);
\coordinate (fork3) at (5.7,-4.8);
\draw[arr] (split.south) -- (fork1);
\draw[arr] (fork1) -- (fork2);
\draw[arr] (fork2) |- (moge.east);
\draw[arr] (fork2) -- (fork3);
\draw[arr] (fork3) -- (qfmt.east);

% HiSpatial row
\draw[arr] (moge.east) -- (webdata.west);
\draw[arr] (webdata.east) -- (htrain.west);
\draw[arr] (htrain.east) -- (eval.north west);

% Qwen row
\draw[arr] (qfmt.east) -- (qlora.west);
\draw[arr] (qlora.east) -- (qtrain.west);
\draw[arr] (qtrain.east) -- (eval.south west);

% Branch labels positioned away from boxes/arrows
\node[text=accent, font=\bfseries\small] at (12.0,-1.45) {HiSpatial branch};
\node[text=accent, font=\bfseries\small] at (12.0,-5.65) {Qwen3-VL branch};
\end{tikzpicture}%
}
\caption{Implemented fine-tuning and evaluation pipeline. The two model-specific branches share the same task supervision and evaluation interface but differ in geometric preprocessing and adaptation strategy.}
\label{fig:pipeline}
\end{figure*}

Figure~\ref{fig:pipeline} summarizes the completed workflow. A shared data layer first prepares model-independent VQA records and validates image--annotation alignment. The HiSpatial branch then augments each image with a dense 3D point map and serializes the result for full-model supervised fine-tuning, whereas the Qwen3-VL branch converts the same task supervision into multimodal conversational instructions for QLoRA adaptation. Both branches converge on a common checkpointed inference and automated evaluation stage.

\subsection{HiSpatial Data Preparation}
The HiSpatial training stack requires synchronized RGB inputs, spatial tensors, and VQA metadata. The implemented preprocessing stage performs four main operations:
\begin{enumerate}
    \item load the target scene image and its aligned VQA records;
    \item run MoGe on the image to produce a dense XYZ tensor;
    \item serialize the image, geometry tensor, and conversation/answer metadata into WebDataset-style samples;
    \item mix target-domain samples with general visual instruction data before fine-tuning.
\end{enumerate}

In the first successful adaptation setting, target-domain building data were mixed with LLaVA-NeXT-style general VQA samples at approximately a 1:1 ratio. Prior work on broad visual instruction tuning highlights the value of diverse supervision for preserving general multimodal behavior~\cite{llavanext}. This consideration is also supported by our experiments: highly concentrated spatial supervision can teach the target answer format effectively while simultaneously narrowing the response distribution of a general VLM.

The prepared HiSpatial samples contain three synchronized components:
\begin{equation}
    \mathcal{D}_{\text{HiSpatial}}=\{(\mathbf{I}_i,\mathbf{P}_i,\mathbf{c}_i)\}_{i=1}^{N},
\end{equation}
where $\mathbf{P}_i$ is the precomputed point map and $\mathbf{c}_i$ is the VQA conversation. Precomputing and serializing the point maps decouples geometry estimation from model optimization, reduces repeated preprocessing cost, and improves experiment reproducibility.

\subsection{HiSpatial Fine-Tuning and Checkpoint Evaluation}
Training and inference are maintained in separate environments to reduce dependency conflicts. Fine-tuning starts from the released HiSpatial-3B checkpoint rather than from random initialization. The initial validated configuration used a per-device batch size of 4, an effective total batch size of 32, 1,200 optimization steps, and checkpointing every 400 steps. Subsequent runs extended the training horizon and adjusted the data mixture while preserving the same end-to-end pipeline.

Training can be launched through the project-level wrapper:
\begin{lstlisting}[style=tech,language=bash]
# HiSpatial training environment
conda activate hispatial-train

# Example launch
bash scripts/run_train.sh configs/train_finetune_lhzzzzzy.json
\end{lstlisting}

Checkpoint evaluation is fully automated. Each selected checkpoint is loaded, evaluated over a JSONL test split, and exported to a standardized prediction file that is then consumed by the task-aware evaluator. The same interface also supports later styled-image variants and colored bounding-box prompts, allowing new visual-input formats to be tested without changing the evaluation protocol.

\subsection{Qwen3-VL Instruction Construction and QLoRA}
For Qwen3-VL, the same VQA supervision is transformed into the multimodal conversational format expected by the fine-tuning framework. A representative training record is
\begin{lstlisting}[style=tech,language={}]
{
  "image": [".../sample_000101/rgb_mask_id_overlay.png"],
  "conversations": [
    {"from": "human",
     "value": "How many buildings are ...?\n<image>"},
    {"from": "gpt",
     "value": "There are ... buildings."}
  ]
}
\end{lstlisting}

The first experiment used 1,000 scenes, with 900 assigned to training and 100 to testing. Because each scene contributes 12 question types, this split yields approximately 10,800 training QA instances and 1,200 test instances.

The implemented Qwen adaptation uses 4-bit quantization with low-rank adapters to reduce memory consumption. In the lightweight recipe, the vision encoder remains frozen while LoRA parameters are optimized in language-side transformer projections; the multimodal connector can also be enabled depending on the configuration. This design makes 8B- and 32B-scale specialization feasible without maintaining gradients for the full pretrained parameter set.

\subsection{Unified Inference and Evaluation}
A common inference interface maps each sample identifier to its image, executes the selected model/checkpoint, and writes one normalized JSON object per prediction. The downstream evaluator then applies task-specific scoring: direct matching for counting, numerical parsing with relative tolerance for metric estimation, and semantic normalization for open-form spatial answers.

A unified multi-model runner additionally allows HiSpatial, Qwen3-VL, LLaVA-3D, and SpatialRGPT to consume the same JSONL benchmark. This abstraction removes model-specific evaluation scripts from the comparison loop and makes checkpoint/model comparisons substantially more reproducible.

\section{Experimental Results and Engineering Findings}
\subsection{HiSpatial Reproduction Sanity Check}
Before target-domain adaptation, the reproduced HiSpatial training and inference stack was validated on standard spatial benchmarks. Despite using substantially less compute than the original training setup, the reproduced pipeline achieved 94.17\% on CV-Bench-3D, 94.77\% on CV-Bench Relation, 73.90\% on EmbSpatial, and 78.05\% on RoboSpatial in the project records. These measurements are used as an implementation sanity check rather than as a claim of full reproduction of the original training regime.

\subsection{Target-Domain HiSpatial Adaptation}
The clearest quantitative outcome of the completed engineering work is the successful target-domain adaptation of HiSpatial. As shown in Table~\ref{tab:hispresults}, overall accuracy on the original building-centric benchmark rises from 1.73\% before adaptation to 64.69\% after 1,200 steps, 73.90\% after 2,400 steps, and 79.89\% after 3,600 steps.

\begin{table}[t]
\centering
\caption{Recorded HiSpatial accuracy on the original target dataset during fine-tuning.}
\label{tab:hispresults}
\small
\begin{tabular}{lc}
\toprule
\textbf{Checkpoint / model} & \textbf{Overall accuracy (\%)} \\
\midrule
Released HiSpatial before adaptation & 1.73 \\
Fine-tuned, 1,200 steps & 64.69 \\
Fine-tuned, 2,400 steps & 73.90 \\
Fine-tuned, 3,600 steps & 79.89 \\
\bottomrule
\end{tabular}
\end{table}

The improvement is not confined to the original target split. On a separately collected styled dataset, accuracy increases from 14.9\% for the released model to 49.2\%, 52.8\%, and 55.1\% for the 1,200-, 2,400-, and 3,600-step checkpoints, respectively. The consistent gain on this separate split provides evidence that the pipeline is learning transferable task structure rather than merely memorizing the original scenes.

A later data-mixture experiment further showed that answer-format composition materially affects adaptation. Naively combining free-form and multiple-choice supervision caused a pronounced drop on the original free-form task, whereas an 8:2 mixture recovered strong performance. This result indicates that output-format distribution should be treated as an explicit component of the training domain rather than as an incidental formatting choice.

\subsection{Qwen3-VL Fine-Tuning with Mixed-Domain Supervision}
The Qwen3-VL adaptation pipeline was implemented for both the 8B and 32B variants using memory-efficient QLoRA. Following the data-mixture strategy adopted for HiSpatial, the target-domain spatial instruction data were combined with general multimodal instruction samples during fine-tuning. This mixed-domain formulation was designed to adapt the models to the target spatial tasks while retaining the broader visual-language capabilities of the pretrained models.

The resulting checkpoints show clear improvements over the corresponding pretrained baselines. As summarized in Table~\ref{tab:qwenresults}, Qwen3-VL-8B improves from 20.74\% before fine-tuning to 45.27\% after mixed-domain adaptation, while Qwen3-VL-32B improves from 30.79\% to 47.60\%. These results demonstrate that the implemented QLoRA pipeline can successfully specialize both model scales to the target spatial benchmark when the target supervision is balanced with more general multimodal instruction data.

\begin{table}[t]
\centering
\caption{Qwen3-VL accuracy before and after mixed-domain QLoRA fine-tuning.}
\label{tab:qwenresults}
\small
\begin{tabular}{lcc}
\toprule
\textbf{Model} & \textbf{Before FT} & \textbf{After FT} \\
\midrule
Qwen3-VL-8B  & 20.74\% & 45.27\% \\
Qwen3-VL-32B & 30.79\% & 47.60\% \\
\bottomrule
\end{tabular}
\end{table}

The improvement is consistent with the behavior observed in the HiSpatial experiments: concentrated target-domain supervision alone is not necessarily the most effective adaptation strategy for a general-purpose multimodal model, whereas mixed-domain training provides a better balance between task specialization and capability retention. In the present implementation, the same general strategy is therefore used for both model families, although the optimization mechanisms differ: HiSpatial is adapted with its RGB+3D training pipeline, while Qwen3-VL is optimized through low-rank adapters on top of the quantized pretrained model.

The successful Qwen3-VL experiments also complete the second model branch of the engineering pipeline. The implementation covers instruction conversion, mixed-domain dataset construction, QLoRA optimization, checkpoint serialization and reloading, standardized batch inference, and automated evaluation for both 8B- and 32B-scale models.

\subsection{Current Implementation Status}
Table~\ref{tab:status} summarizes the completed components.

\begin{table}[t]
\centering
\caption{Current engineering status.}
\label{tab:status}
\small
\begin{tabular}{p{0.68\columnwidth} c}
\toprule
\textbf{Component} & \textbf{Status} \\
\midrule
Dataset JSONL conversion and image alignment & Complete \\
Bounding-box / styled-image input support & Complete \\
MoGe point-map preprocessing & Complete \\
HiSpatial WebDataset construction & Complete \\
HiSpatial mixed-domain full fine-tuning & Complete \\
HiSpatial checkpoint inference/evaluation & Complete \\
Qwen3-VL instruction conversion & Complete \\
Qwen3-VL mixed-domain QLoRA fine-tuning & Complete \\
Unified prediction and automated evaluation & Complete \\
8B/32B checkpoint evaluation & Complete \\
\bottomrule
\end{tabular}
\end{table}

\section{Reproducibility and Software Organization}
The software stack separates data conversion, training, inference, and evaluation environments in order to minimize dependency conflicts across model families. The principal environments are \texttt{hispatial}, \texttt{hispatial-train}, and \texttt{qwen3vl}, with additional environments reserved for spatial-VLM baselines.

Each experiment is designed to be recoverable from four artifacts: (1) the dataset JSONL, (2) the corresponding image root, (3) the model checkpoint and configuration, and (4) the exported evaluation results. A minimal HiSpatial evaluation run therefore has the form
\begin{lstlisting}[style=tech,language=bash]
conda activate hispatial-train
python tools/run_hispatial_checkpoint_jsonl.py \
  --dataset <test.jsonl> \
  --image-root <image_dir> \
  --checkpoint <checkpoint.ckpt> \
  --output <predictions.jsonl>

conda activate hispatial
python tools/eval_data4_color_preds.py \
  --input <predictions_aug.jsonl> \
  --output <evaluation.json>
\end{lstlisting}

The same convention is used for Qwen3-VL: each checkpoint is passed through the standardized runner and produces predictions in the same JSONL schema. Consequently, downstream model comparison does not require evaluator-specific changes. The use of a shared data interface and evaluation protocol also makes it possible to compare different adaptation strategies without modifying the benchmark implementation.

\section{Discussion and Next Steps}
The experiments lead to three main observations. First, the HiSpatial pipeline performs strongly on the target spatial benchmark, benefiting from explicit RGB+point-map inputs for metric reasoning. Second, mixed-domain training is effective for both model families. Under this strategy, Qwen3-VL-8B improves from 20.74\% to 45.27\%, while Qwen3-VL-32B improves from 30.79\% to 47.60\%. Third, HiSpatial and Qwen3-VL provide complementary adaptation strategies: the former relies on explicit 3D geometry, whereas the latter achieves substantial gains through parameter-efficient multimodal instruction tuning.

Future work will focus on target/general data-mixture ratios, task-wise evaluation, input-format ablations, and comparisons between monocular point maps and geometry derived directly from the available 3D assets or camera metadata.

\section{Conclusion}
This work establishes two end-to-end adaptation pipelines for a custom 3D-aware spatial benchmark. The HiSpatial pipeline improves overall accuracy from 1.73\% to 79.89\% and raises performance on a separately collected styled split from 14.9\% to 55.1\%. The Qwen3-VL pipeline is implemented for both 8B and 32B models using mixed-domain QLoRA, improving accuracy from 20.74\% to 45.27\% and from 30.79\% to 47.60\%, respectively.

Overall, the project provides a reproducible engineering foundation for further study of metric 3D reasoning, spatial grounding, multimodal data mixing, and geometry-aware model adaptation.

\balance


\begin{thebibliography}{99}

\bibitem{hispatial}
H. Liang, Y. Shen, Y. Deng, S. Xu, Z. Feng, T. Zhang, Y. Liang, and J. Yang,
``HiSpatial: Taming Hierarchical 3D Spatial Understanding in Vision-Language Models,''
\emph{Proc. IEEE/CVF Conf. Computer Vision and Pattern Recognition (CVPR)}, 2026.

\bibitem{qwen3vl}
S. Bai, Y. Cai, R. Chen, \emph{et al.},
``Qwen3-VL Technical Report,''
\emph{arXiv preprint arXiv:2511.21631}, 2025.

\bibitem{moge}
R. Wang, S. Xu, C. Dai, J. Xiang, Y. Deng, X. Tong, and J. Yang,
``MoGe: Unlocking Accurate Monocular Geometry Estimation for Open-Domain Images with Optimal Training Supervision,''
\emph{arXiv preprint arXiv:2410.19115}, 2024.

\bibitem{paligemma2}
A. Steiner, A. S. Pinto, M. Tschannen, \emph{et al.},
``PaliGemma 2: A Family of Versatile VLMs for Transfer,''
\emph{arXiv preprint arXiv:2412.03555}, 2024.

\bibitem{qlora}
T. Dettmers, A. Pagnoni, A. Holtzman, and L. Zettlemoyer,
``QLoRA: Efficient Finetuning of Quantized LLMs,''
\emph{Advances in Neural Information Processing Systems (NeurIPS)}, 2023.

\bibitem{llavanext}
F. Li, R. Zhang, H. Zhang, Y. Zhang, B. Li, W. Li, Z. Ma, and C. Li,
``LLaVA-NeXT-Interleave: Tackling Multi-image, Video, and 3D in Large Multimodal Models,''
\emph{arXiv preprint arXiv:2407.07895}, 2024.

\end{thebibliography}
\end{document}